\documentclass[preprint, 1p]{elsarticle}

\makeatletter
\def\ps@pprintTitle{%
  \let\@oddhead\@empty
  \let\@evenhead\@empty
  \let\@oddfoot\@empty
  \let\@evenfoot\@oddfoot
}
\makeatother

\usepackage{times}
\usepackage{epsfig}
\usepackage{graphicx}
\usepackage{amsmath}
\usepackage{amssymb}
\usepackage{booktabs}
\usepackage{wrapfig}
\usepackage{subcaption}
\usepackage[pagebackref=true,breaklinks=true,colorlinks,bookmarks=false]{hyperref}
\usepackage[dvipsnames]{xcolor}
\usepackage[inline]{enumitem}

\graphicspath{{./images/}}

\newtheorem{defn0}{Definition}[section]
\newtheorem{prop0}[defn0]{Proposition}
\newtheorem{thm0}[defn0]{Theorem}
\newtheorem{lemma0}[defn0]{Lemma}
\newtheorem{corollary0}[defn0]{Corollary}
\newtheorem{example0}[defn0]{Example}
\newtheorem{remark0}[defn0]{Remark}
\newtheorem{conjecture0}[defn0]{Conjecture}

\DeclareMathOperator*{\argmax}{arg\,max}

\numberwithin{equation}{section}
\numberwithin{figure}{section}

\begin{document}

\title{Predicting the generalization gap in neural networks using topological data analysis}

\author[1,2]{Rub\'en Ballester \corref{cor1}}
\ead{ruben.ballester@ub.edu}
\author[1]{Xavier Arnal Clemente}
\ead{xavi.aclm@gmail.com}
\author[1]{Carles Casacuberta
\fnref{fn1}}
\ead{carles.casacuberta@ub.edu}
\author[2]{Meysam Madadi}
\ead{meysam.madadi@gmail.com}
\author[3]{Ciprian~A. Corneanu}
\ead{cipriancorneanu@gmail.com}
\author[1,2]{Sergio Escalera
\fnref{fn2}}
\ead{sergio.escalera.guerrero@gmail.com}
\address[1]{Departament de Matem\`atiques i Inform\`atica, Universitat de Barcelona, Gran Via de les Corts Catalanes 585, 08007~Barcelona, Spain}
\address[2]{Computer Vision Center, Campus UAB, 08193~Bellaterra, Barcelona, Spain}
\address[3]{Amazon Apollo, 25 9th Ave N, Seattle, WA 98109, United States}

\cortext[cor1]{Corresponding author}
\fntext[fn1]{Partially supported by MCIN/AEI/10.13039/501100011033 under grant PID2020-117971GB-C22}
\fntext[fn2]{Partially supported by the Spanish project PID2022-136436NB-I00 and by the ICREA Academia programme}

\begin{abstract}
    Understanding how neural networks generalize on unseen data is crucial for designing more robust and reliable models. In this paper, we study the generalization gap of neural networks using methods from topological data analysis.
    For this purpose, we 
    compute homological persistence diagrams of weighted graphs 
    constructed from neuron activation correlations after a training phase, 
    aiming to capture patterns that are linked to the generalization capacity of the network.
    We compare the usefulness of different numerical summaries from persistence diagrams and show that a combination of some of them can accurately predict and partially explain the generalization gap without the need of a test set. Evaluation on two computer vision recognition tasks (CIFAR10 and SVHN) shows competitive generalization gap prediction when compared against state-of-the-art methods.
\end{abstract}

\begin{keyword}
Deep learning\sep neural network\sep topological data analysis\sep generalization gap
\end{keyword}

\maketitle

\section{Introduction}

\noindent
Understanding the generalization capacity of a neural network is one of the most important questions in deep learning. Unfortunately, while the fundamental procedures of training neural networks are well understood, being able to tell why one network is better at generalizing than another still poses a great challenge. Good performance of a deep neural network (DNN) depends fundamentally on its architecture and its neuron functions and parameters. These yield an approximation of the desired function (prediction or regression) based on neuron interactions ---the better the approximation, the better the generalization. However, with the high quantity of neurons and connections of deep neural networks (sometimes of the order of millions), understanding which interactions between neurons are improving or damaging a model is a hard problem. 
Developing new mathematical tools that capture the effect of these interactions on the output of the networks is key for increased understanding of network generalization.

A DNN that generalizes will perform well on test data on which it has not been trained. This is usually measured by the generalization gap, which is defined as the difference between the accuracy in training vs.\ test datasets. Although the two accuracies are correlated to a certain extent, studying training performance alone can be misleading. Several papers show how neural network performances on unseen examples can differ with respect to their training performances due to many reasons
\cite{azulay2019deep, goodfellow2015explaining, zhang2017understanding}.
\emph{To what extent is it possible to predict the generalization gap without testing a model?} In a practical sense, a measure of generalization that does not require a test dataset eliminates the responsibility of maintaining and curating such a dataset. 

The issue of finding a generalization measure has been explored extensively and a recent challenge on the topic provides an excellent framework for algorithmic benchmarking \cite{generalizationChallenge}. However, the most competitive participant methods rely on internal representations of independent layers, discarding more global structures that may be created across the network \cite{brain,interpex} or even discarding structure altogether \cite{alwaysGen}.

An alternative approach is provided by topological data analysis (TDA), an applied branch of algebraic topology that studies the shape of sets of points endowed with a metric structure. Such shapes are described by means of persistence diagrams \cite{COMPTOP}, which are built on homological features of simplicial complexes constructed from the given dataset.

In this paper we present an approach to predict the generalization gap from persistence diagrams based on neuron interactions in deep neural networks of any size.
For this purpose, we use
weighted graphs 
computed from activation correlations between neurons after training a network with a dataset.
We compare the performance of different persistence diagram vectorizations, called persistence summaries, from which the generalization gap can be regressed, and we find that a suitable combination of such summaries 
yields competitive results on measuring the generalization~gap. 
Moreover, we show that persistence summaries separate neural network architectures into clusters related with their generalization capacity.

The paper is structured as follows: in Section~\ref{sec:related_work} we discuss related work; in Section~\ref{sec:methodology} we define functional graphs and describe their persistence summaries; in Section~\ref{sec:results} we present and discuss experimental results, and conclusions are written down in Section~\ref{sec:conclusions}. 
Supplemental material is provided in an appendix. \let\thefootnote\relax\footnotetext{\hspace*{-0.5cm}The code for this article is available in the following repository: \newline \url{https://github.com/rballeba/PredictingGeneralizationGapUsingPersistentHomology}}

\section{Related Work}
\label{sec:related_work}

\noindent \textbf{Predicting generalization.} Understanding the generalization gap is a major area of research in theoretical and practical deep learning. One of the most influential papers in the last few years has been~\cite{understanding_deep_learning}, in which classical theories on the generalization capabilities of machine learning models were shown to fail to explain why neural networks generalize well in practice. This paper motivated a tremendous amount of original work on generalization of deep neural networks. From the theoretical point of view, some works tried to correct the flaws of the previous methods by developing new and tighter generalization bounds~\cite{dziugaite2017computing, nagarajan2019generalization, Kawaguchi_2022, NIPS2017_b22b257a, pmlr-v75-golowich18a, pmlr-v89-liang19a, dupuis2023generalization, lotfi2022pacbayes, NEURIPS2020_37693cfc, hardt2015train}, by studying generalization measures~\cite{NIPS2017_10ce03a1, jiang2019predicting}, or by studying the training process~\cite{AshiaMarginal, chaudhari2018stochastic, smith2017bayesian}, among others. From an experimental point of view, there have been many works studying generalization measures and trying to predict the generalization gap. 
One of the most extensive benchmarks for the robustness of generalization measures was developed in~\cite{jiang2019fantastic}, where $40$ different generalization measures were tested in more than $10{,}000$ trained models. With the objective of developing new robust generalization measures, the first competition on predicting generalization in deep learning (PGDL) was organized at NeurIPS~\cite{generalizationChallenge} and its results were published in~\cite{pmlr-v133-jiang21a}. The generalization measures presented there were divided into three main categories: 
\begin{enumerate*}
    \item Measures based on theoretical generalization bounds;
    \item Measures based on data augmentation; and
    \item Measures based on the analysis of intermediate representations.
\end{enumerate*}
The winners of the competition, the teams Interpex~\cite{interpex}, Always Generalize~\cite{alwaysGen}, and BrAIn~\cite{brain}, presented generalization measures in the last two categories. The Interpex team proposed a generalization measure based on neuroscience ideas that uses the Davies--Bouldin index~\cite{4766909} to quantify the consistency of internal representations of neural networks, the Always Generalize team proposed to measure the robustness of neural networks against data-augmented datasets, and the BrAIn team proposed a measure based on properties of a graph constructed from the internal representations of a neural network. After the competition, other robust generalization measures were published~\cite{NEURIPS2021_b0dd033c, 9747136}.

The lack of winners based on theoretical generalization bounds suggests that theoretical bounds are still far from being usable in practical scenarios, and that new and original methods are needed to keep improving our understanding of the generalization phenomenon in neural networks. On this aspect, our approach, while being novel in its methodology, obtains state-of-the-art performance when predicting the generalization gap compared with the winning methods of the PGDL competition.

\bigskip

\noindent
\textbf{Topological data analysis.} Topological data analysis (TDA) has been used very successfully in machine learning. 
A survey of applications is offered in~\cite{survey_tml}. 
From a theoretical point of view, topological data analysis has been used to analyze structural properties of neural networks~\cite{8999174}, input and output spaces~\cite{6697897, guss2018characterizing, pmlr-v97-ramamurthy19a,NEURIPS2020_5f146156, 8578494, geometrical_and_topological_properties_of_dnns, petri2020on}, generative models and their properties~\cite{pmlr-v80-khrulkov18a, zhou2021evaluating}, and internal representations and weights of neural networks~\cite{topological_approaches_to_deep_learning, exposition_and_interpretation_of_the_topology_of_neural_networks, neural_persistence, characterising_the_shape_of_activation_space_in_deep_neural_networks, caveats_of_neural_persistence_in_deep_neural_networks}, among others.

In the intersection of topological data analysis with prediction of the generalization gap, we find~\cite{birdal2021intrinsic, whatdoesitmean, testingErrorWithoutDataset}. In \cite{birdal2021intrinsic}, a novel connection between the upper box dimension and the persistent homology dimension~\cite{kozma2006minimal, Schweinhart2021} is used to bound the generalization gap of neural networks using the fractal dimension of training trajectories~\cite{NEURIPS2020_37693cfc}. In \cite{whatdoesitmean, testingErrorWithoutDataset}, generalization of neural networks is studied by calculating persistent homology of activation vectors of the neural network on the training dataset. In particular, in \cite{testingErrorWithoutDataset}, the generalization gap is predicted with linear models based on persistence summaries extracted from neuron activations.

However, the existing methods fail to be suitable in certain scenarios. On the one hand,~\cite{birdal2021intrinsic} cannot be used without the training information of a neural network, which is generally not available when using pretrained models. On the other hand, the methods from~\cite{whatdoesitmean, testingErrorWithoutDataset} 
do not scale to modern neural network architectures, as they compute descriptors from persistence diagrams, which share in most cases a computational complexity higher than cubical on the number of neurons in the network. In addition, the summaries of persistence diagrams tested in these articles are scarce, and other persistence summaries could potentially be better suited to predict the generalization gap.

\subsection{Contributions}

\noindent
In this article, the following contributions are made:
\begin{enumerate}
    \item We extend the methodology of~\cite{whatdoesitmean, testingErrorWithoutDataset} to make it capable of processing large neural networks. To achieve this, we propose a methodology that performs bootstrapping on persistence summaries computed from persistence diagrams coming from different samples of neurons from the same network. Samples are taken following a probability distribution over the set of neurons of the network, giving more probability to neurons that share high activation values.
    \item We train linear models using eleven different combinations of persistence summaries to predict the generalization gap and compare these models with linear models trained on the generalization measures proposed by the PGDL competition winners~\cite{generalizationChallenge}, obtaining competitive results. We find that basic statistical parameters of the distribution of points in persistence diagrams are the best performing summaries to predict generalization gaps.
    \item We offer an interpretation of why topological data analysis is meaningful for predicting the generalization gap from features learned by a neural network. Figure~\ref{fig:avgstdFeature} illustrates a neat clustering phenomenon of network architectures with respect to their depth when the generalization gap is represented in relation with suitable persistence summaries.
\end{enumerate}

\section{Methodology}
\label{sec:methodology}

\noindent 
Let $N\colon \mathcal{X}\to\mathcal{Y}$ denote a classification neural network, where $\mathcal{X}$ is a set of inputs and $\mathcal{Y}$ is a set of labels. Let $\mathcal{L}$ be a loss function on $\mathcal{Y}\times\mathcal{Y}$ that measures the error of a prediction, and let $\mathcal{D}, \mathcal{T}\subseteq \mathcal{X}\times\mathcal{Y}$ be a pair of training and test datasets, respectively. 
Let 
\[
\mathcal{R}[N]= \mathbb E_{(x,y)\sim \mathbb P_{(X,Y)}}[\mathcal{L}(N(x), y)]
\]
be the expected risk of~$N$,
where $\mathbb P_{(X,Y)}$ 
is a generally unknown data distribution, and let
\[
\mathcal R_{S}[N]=
\frac{1}{|S|}
\sum_{i=1}^{|S|} \mathcal{L}(N(x_i), y_i)
\]
be the empirical risk function on a dataset $S=\{(x_i,y_i)\}\subseteq\mathcal{X}\times\mathcal{Y}$.

A main objective in classification tasks is to find an optimal network $N_\text{opt}$ 
from a given set of neural networks that minimizes the expected risk $\mathcal{R}[N]$.
In most cases, $\mathcal{R}[N]$ cannot be computed, since the data distribution function $\mathbb P_{(X,Y)}$ is not known.
Therefore, the usual approach is to minimize the empirical risk function $\mathcal{R}_\mathcal{D}[N]$ using the training dataset $\mathcal{D}$.

In the special case of the 0-1 loss function
$\mathcal{L}(\hat y,y)=1$ if $\hat y=y$ and $0$ otherwise, the empirical risk can be written as 
$\mathcal{R}_\mathcal{D}[N] = 1-\text{Acc}_\mathcal{D}[N]$, 
where $\text{Acc}_\mathcal{D}[N]$ is the training accuracy used as benchmarking measure in most deep learning classification problems. Therefore, minimizing the empirical risk for this function $\mathcal{L}$ 
is equivalent to maximizing the training accuracy. 

However, minimization of empirical risks does not necessarily lead to minimization of expected risks, due to phenomena such as overfitting. The difference 
$\mathcal{R}[N] - \mathcal{R}_\mathcal{D}[N]$
between both quantities is known as the \emph{generalization gap} of the neural network~$N$.
This quantity is usually approximated with the \emph{empirical generalization gap}, which is defined as the difference $\mathcal{R}_\mathcal{T}[N]-\mathcal{R}_\mathcal{D}[N]$ between the empirical risks for the training and test datasets.
For the 0-1 loss function, the empirical generalization gap is equal to the difference $\text{Acc}_\mathcal{D}[N] - \text{Acc}_\mathcal{T}[N]$ between the accuracies in train and in test.
With the realistic assumption that current neural networks obtain better training accuracy than test accuracy and that training accuracies are generally high, a lower generalization gap is an indication of a better network performance.

\subsection{Objectives}
\label{sec:objectives}

\noindent
The main purpose of this paper is to predict the empirical generalization gap using only information from the training dataset $\mathcal{D}$ by gleaming information about the dynamic behaviour of a trained neural network, i.e., the internal representations, structures and relationships between neuron activations during classification. In our context, the network behaves dynamically only in the presence of input data, forming a graph of neuron activations. 

Our first goal is to define a mathematical structure describing the activation of a network when fed with a specific dataset $\mathcal{D}$ 
consisting of pairs $(x,y)$ where $x$ and $y$ represent inputs and ground truth annotations respectively. To do so, we use a complete weighted graph whose set of vertices is in bijective correspondence with the set of neurons of the given network. 
Each vertex in this graph is represented by an activation vector of dimension $\left|\mathcal{D}\right|$ where the vector components are the neuron's activations for all $(x,y)\in\mathcal{D}$. Edges are weighted by a 
correlation distance between the activation vectors that they are connecting.

From this weighted graph
we build a filtered simplicial complex computed from the edge weights, whose topological features are described by a persistence diagram, from which we extract suitable summaries with the purpose of relating them with the empirical generalization gap of the network. Precise definitions are given in the next subsections.

\subsection{Network functional graphs}
\label{sec:methodology:ciprian_spaces}

\noindent
Let $V=\{v_1,\dots,v_n\}$ be the set of non-input nodes of a neural network $N$ trained with a dataset $\mathcal{D}=\{(x,y)\}$, where $x$ denotes inputs and $y$ denotes corresponding values from a set of labels.
For a node $v\in V$, we denote by $N_v(x)$ the activation value of $v$ on some input~$x$, and define the \emph{activation vector} of $v$ as
\begin{equation*}
    A_v(\mathcal{D}) = (N_v(x))_{(x,y)\in\mathcal{D}}.
\end{equation*}
The set $A_N(\mathcal{D}) = \{A_v(\mathcal{D})\mid v\in V\}$ of activation vectors is meant to capture the role of each node of $N$ during inference.

A \emph{correlation distance} between two nodes $v_i,v_j\in V$ is defined as
\begin{equation}
\label{distance}
   d(v_i,v_j) = 
   1-|\text{corr}(A_{v_i}(\mathcal{D}),A_{v_j}(\mathcal{D}))|, 
\end{equation}
where $\text{corr}$ is the Pearson correlation coefficient.
Nodes with constant activations can be safely regarded as not affecting the behaviour of the model, but rather its structure as a bias. Therefore, nodes with zero variance are discarded.
Although this function $d$ does not satisfy the axioms of a metric, it is suitable for the application of techniques from TDA
---this fact is discussed in Section~\ref{correlation} below.

The complete weighted graph with vertices the nodes in $V$ with nonzero variance and weights $d(v_i,v_j)$ on the edges will be called the \emph{functional graph} of the trained neural network~$N$.
This graph encodes the functional behaviour of~$N$.
In this article we use Vietoris--Rips filtrations associated with the distance matrix $(d(v_i,v_j))$ from the functional graph for a homological persistence study, as defined in the next section.

\subsection{Topological Data Analysis}

\subsubsection{Vietoris--Rips complexes}

\noindent
An abstract simplicial complex, a basic tool of algebraic topology, is a finite collection of sets $S$ such that if $\alpha \in S$ and $\beta \subseteq \alpha$ then $\beta \in S$. Each abstract simplicial complex $K$ determines a sequence of \emph{homology groups} $H_n(K)$ for $n\ge 0$, 
generated by linearly independent $n$-dimensional cycles modulo boundaries. In this article coefficients of homology groups are meant in the field $\mathbb{F}_2$ of two elements.

If $V$ is a finite set equipped with a distance function~$d$, then for each subset $\alpha\subseteq V$ we may consider the diameter
$\text{diam}(\alpha) = \max_{i, j\in{\alpha}}{d(i,j)}$
of $\alpha$ relative to~$d$. 
The \emph{Vietoris--Rips complex} of $V$ at a parameter value $r\geq 0$ is an abstract simplicial complex defined as 
\begin{equation*}
    \text{VR}_r(V) = \{\alpha\subseteq V: \text{diam}(\alpha)\leq{r}\}.
\end{equation*}

The set $\{\text{VR}_r(V)\}_{r\ge 0}$ is a nested collection of simplicial complexes,
as $\text{VR}_r(V)\subseteq \text{VR}_s(V)$ if $r\le s$.
Each such filtration yields a \emph{persistence diagram} for every integer~$n\geq 0$, which contains a point $(r,s)$ for each homology generator of dimension $n$ born at a parameter value $r$ and vanishing at~$s$, where $r<s$. Further details about persistence diagrams can be found in~\cite{COMPTOP}.

\subsubsection{Correlation distance}
\label{correlation}

\noindent
The correlation distance $d$ defined in \eqref{distance} can take a zero value on distinct nodes and the triangle inequality need not hold. However, Vietoris--Rips filtrations can be associated with arbitrary functions $X\times X\to\mathbb{R}$ where $X$ is any set, and stability holds in such generality \cite{cso2014,cm2018}.

Although $d$ does not necessarily satisfy that 
$d(x,y)\neq 0$ whenever $x\neq y$,
this does not affect  persistent homology, since the matrix $(d(v_i,v_j))$ yields 
Vietoris--Rips complexes
homotopy equivalent to those obtained by identifiying two nodes $x$ and $y$ if $d(x,y)=0$.
Moreover, while $d$ does not satisfy the triangle inequality, 
the following transformation does:
\[
d'(v_i,v_j)=\sqrt{1-(1-d(v_i,v_j))^2}.
\]
Since the function $\gamma(t)=\sqrt{1-(1-t)^2}$ is monotonic on $[0,1]$ and uniformly continuous, $d$ and $d'$ produce the same Vietoris--Rips filtrations, albeit at different thresholds, and share similar continuity properties with respect to small displacements in the space of functional graphs. 

\subsubsection{Persistence summaries}
\label{methodTopoSum}

\noindent
There is a variety of numerical or vector-valued functions defined on persistence diagrams available for statistical analyses. We refer to such functions as \emph{persistence summaries} or \emph{descriptors}.
In this subsection we present the summaries that have been used in our work. 

\medskip

\noindent
\textbf{Average and standard deviation of lifetime parameters.}
\label{logModel}
Different combinations of birth parameters and death parameters have been explored in this article, including their squares and the transformation $1/x + \ln x$ applied element-wise.
We used averages and standard deviations of births and deaths as predictors of the generalization capacity of a network.

The \emph{life} or \emph{lifetime} of a point $(b,d)$ in a persistence diagram is defined as $d-b$,
while the \emph{midlife} is $(b+d)/2$.
Average lives and average midlives also yield useful results when predicting generalization gap using linear extrapolations; these summaries have been used previously with a similar purpose in \cite{testingErrorWithoutDataset}. 
Standard deviation or variance of lives and midlives work equally well or better.
This technique is based on the heuristic that the generalization gap of a network is influenced by the average position and dispersion of  points in persistence diagrams.

\medskip

\noindent
\textbf{Persistence entropy.} The definition of persistence entropy is an adaptation of the concept of entropy used in information theory, which, according to \cite{infoTheory}, provides a measure of the uncertainty of some random variable. The \emph{entropy} of a persistence diagram $P$ is defined as 
\begin{equation}
\label{entropy}
    \epsilon(P)=-\!\!\!\sum_{(b,d)\in{P}}\,((d-b)/L)\, \log_2((d-b)/L)), 
\end{equation}
where 
$L=\sum_{(b,d)\in{P}}{(d-b)}$.
If one defines a discrete random variable that picks points $(b,d)$ from $P$ weighted according to their life, then the persistence entropy corresponds to the entropy of this random variable.
This choice of weights is based on the assumption that points near the diagonal carry less information. More details on persistence entropy can be found in~\cite{pentropy}.

\medskip

\noindent
\textbf{Persistence pooling vectors.}
Persistence pooling vectors were introduced in \cite{persistencePooling} in order to improve a max-pooling procedure using TDA.
This approach consists of analyzing only the most important points in a given persistence diagram, where importance is weighted according to the difference $d-b$. We define the $n$-th persistence pooling vector as the vector in descending order of the $n$ maximum life values. If the persistence diagram has less than $n$ points, then the void vector components are set to~$0$.
We selected the  highest $10$ life values. This number has been chosen experimentally in view of the lack of score performance observed when selecting a larger number of vector components.

\medskip

\noindent
\textbf{Complex polynomials.}
The persistence summary introduced in \cite{complexpol} transforms persistence diagrams into polynomials with coefficients in the field $\mathbb{C}$ of complex numbers whose roots are the images of persistence diagram points under a well-chosen mapping from $\mathbb{R}^2$ to $\mathbb{C}$.
In our study we used the transformation $T$ defined in \cite{complexpol}.

\section{Results}
\label{sec:results}

\noindent
In the first part of this section, we describe experimental setups and comment on computational complexity (\ref{sec:experimentsMain}). In the second part, we evaluate our approach and discuss results (\ref{sec:disc}).

\subsection{Experiments}
\label{sec:experimentsMain}
\begin{figure*}[ht]
\centering
\includegraphics[width=0.99\textwidth]{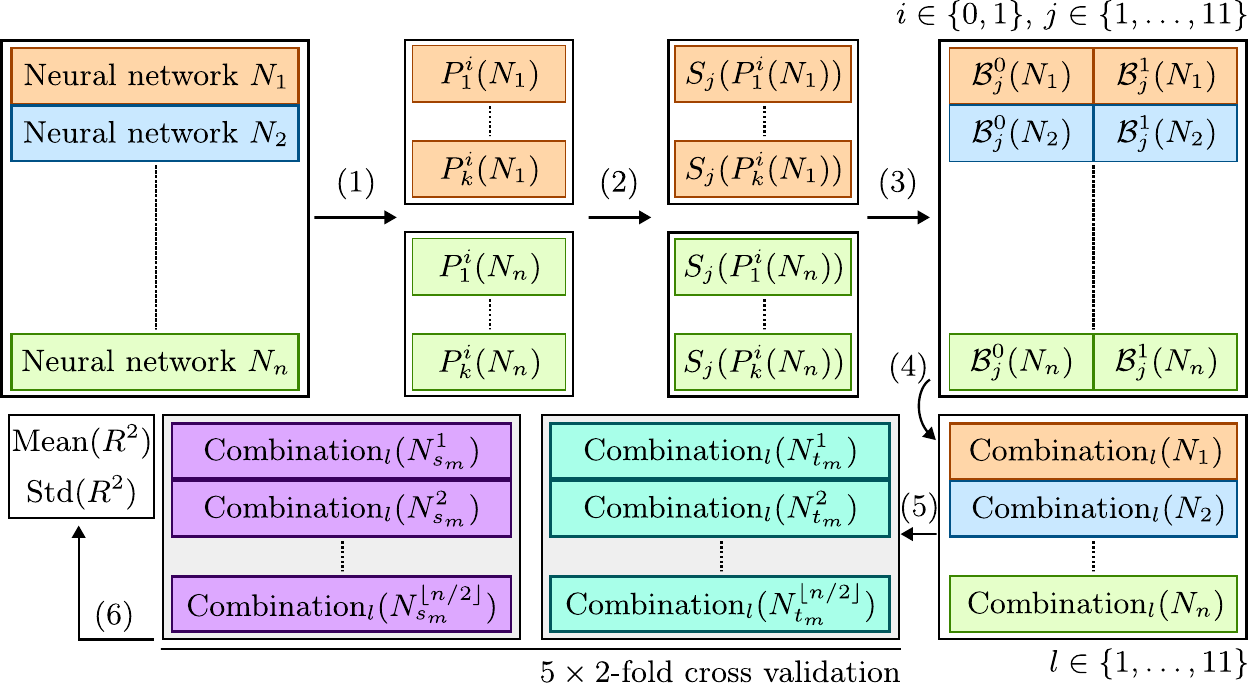}
\caption{Experimental evaluation pipeline. 
Given a specific PGDL task as described in Section~\ref{sec:experimentsMain}, let $\{N_i\}_{i=1}^n$ be the set of neural networks associated with the task.
(1)~Generation of $k$ different persistence diagrams per DNN and dimension $i\in\{0,1\}$ using sampling in CIFAR10/SVHN datasets as described in Section~\ref{datasampling1}. In our case, $k=20$. (2)~Computation of persistence summaries $S_j$ introduced in Section~\ref{methodTopoSum} for each persistence diagram. 
(3)~Bootstrapping for each dimension and each summary computed from the same DNN. The bootstrapped summary $S_j$ for dimension $i$ and neural network $N$ is denoted by $\mathcal{B}_j^i(N)$. (4)~Generation of the eleven different combinations of bootstrapped persistence summaries described in the experimental procedure of Section~\ref{sec:experimentsMain}. 
(5)~A $2$-fold cross-validation partition into sets with the same cardinality is calculated five times. Each time, for each combination of summaries $l\in\{1,\hdots,11\}$, two linear models to predict the generalization gap are trained on one of the partition sets and tested on the other, obtaining a $R^2$ score for each model on the test set. (6)~We compute the mean and standard deviation of the resulting $R^2$ scores. Next, using the same partition sets, we train linear models with the generalization measures of the three winners of the PGDL competition, and we compare our best performing methods with their methods using $5\times 2$-fold cross-validation statistical tests.}
\label{fig:pipelineIsolated}
\end{figure*}

\medskip

\noindent
\textbf{Datasets.}
\label{sec:experimentalSetup}
We use the dataset of trained DNNs provided by the \textit{Predicting Generalization In Deep Learning} (PGDL) competition~\cite{generalizationChallenge}. The dataset is divided into eight tasks, each composed of several neural network architectures trained to provide different generalization gaps on a particular dataset. We focus on the first two tasks, which were public when the competition was launched. 
The first task consists of 96 VGG-like~\cite{Simonyan15} neural networks, with a varying number of dense and convolutional layers (that is, between 2 and 6 per layer type), trained on the CIFAR10 dataset~\cite{cifar}. The CIFAR10 dataset consists of 60{,}000 $32\times 32$~color images (3~channels) in 10~classes, representing vehicles (airplanes, automobiles, ships and trucks) and animals (birds, cats, deers, dogs, frogs, and horses). The second task is composed of 54 neural networks with \textit{network in network} architectures \cite{lin2014network}, with a varying number of blocks, trained on the SVHN dataset \cite{SVHN}. The SVHN is a digit classification benchmark dataset that contains 600{,}000 $32\times 3$2 color images (3~channels) of printed digits (from 0 to 9, 10~classes) cropped from pictures of house number plates. 

\medskip

\noindent
\textbf{Experimental procedure.}
Our experimental procedure is illustrated in Fig.~\ref{fig:pipelineIsolated}. First, we generate $20$ distinct persistence diagrams of dimensions zero and one for each neural network using the sampling methods described in Section~\ref{datasampling1}. After this, we compute
for each persistence diagram the persistence summaries introduced in Section~\ref{methodTopoSum}, generating $20$ different instances of the persistence summary for each network and homology dimension. Then, we compute bootstrapped persistence summaries on each group of $20$ persistence summaries, extracted from the same network, homology dimension, and persistence summary. The bootstrapping process is carried out with $1{,}000$ bootstrap samples of size $20$ taken with replacement over all the different persistence summaries. We combine 
bootstrapped persistence summaries to use them as predictor variables of the generalization gap with a 
linear regression for both tasks. 
The list of persistence summaries that we test is the following:
(1)~Persistence pooling of 10 elements;
(2)~average lives and average midlives;
(3)~average births and average deaths;
(4)~average and standard deviation of births and deaths;
(5)~persistence entropy;
(6)~complex polynomials with 10 coefficients.

We also test concatenations of (2), (3) and (4) with their element-wise squared versions, and a concatenation of (3) with its element-wise logarithmic version, as well as a concatenation of (2) and~(3), original and squared.
All combinations are considered in homological dimension zero, homological dimension one, and a concatenation of both.

We compare our models with linear regressions trained from the three generalization measures that won the PGDL competition; see Section~\ref{sec:related_work} for further details on these generalization measures.

\medskip

\noindent
\textbf{Evaluation metrics.}
\label{sec:evMetrics}
We train linear regression models with the previous combinations of persistence summaries and the three state-of-the-art generalization measures to predict the generalization gap of neural networks. To measure and compare their performance, we use a $5\times 2$-fold cross-validation statistical test, as recommended in~\cite{statistical_tests}, with the coefficient of determination $R^2$ as performance metric. 
The coefficient of determination $R^2$ is computed as the proportion of the variation in the dependent variable that can be predicted from the independent variables, and it is calculated as
\begin{equation}
    \label{rsquare}
R^2(y, \hat{y}) = 1 - \frac{\sum_{i=1}^{n} (y_i - \hat{y}_i)^2}{\sum_{i=1}^{n} (y_i - \bar{y})^2},
\end{equation}
where $y$ denotes the ordered set of actual values, $\hat{y}$ denotes the ordered set of predicted values, and $\bar{y}$ denotes the mean of~$y$.
This coefficient ranges from $0$ to $1$ in the training dataset but can be outside that range in unseen data.
When the score is~$1$, the model perfectly predicts the values of~$y$. 
A score of $R^2=0$ is obtained when one uses a horizontal line at the average of the set of $y$-values as a model.
If a model performs worse than this (which usually indicates that the choice of model itself was ill-advised), then the numerator of \eqref{rsquare} can grow arbitrarily large, and thus $R^2$ can be negative.
If an $R^2$ value is negative, then the prediction is worse than ignoring the input and predicting the average of the sample. 
This can actually happen when the training set yields a model that does not generalize in the test set.

The $5\times 2$-fold cross-validation statistical test validates if there are significant differences between two models tested in a common dataset. 
The null hypothesis of this test is that, for a fixed-size random drawn training dataset, two learning algorithms have the same $R^2$ score on a randomly drawn test dataset. We compare linear models pairwise for each task.

\subsubsection{Reducing computational complexity}
\label{datasampling1}

\medskip
\noindent
\textbf{Computational complexity.} 
    Computing topological summaries with the complete set of activations calculated from the entire training dataset is unfeasible due to the high computational time and memory complexities of obtaining activation vectors and persistence diagrams. If $|\mathcal{D}|$ denotes the number of input samples for a dataset $\mathcal{D}$ and $|V|$ is the number of nodes in a neural network $N$, then the set of activation vectors of nodes in $N$ for the dataset $\mathcal{D}$ has cardinality $|A_N(\mathcal{D})| = |\mathcal{D}| \times |V|$ (see Section \ref{sec:methodology:ciprian_spaces} for details). Assuming that we have a standard current neural network like VGG16, that has about $8{,}000$ neurons~\cite{zhou2020improving} only for fully connected layers, a standard dataset like CIFAR10~\cite{cifar} with $50{,}000$ training examples, and a double precision floating point format to represent each number, one would need at least $3$~GB only to store the activations of fully connected layers. Additionally,
although zero dimensional persistent homology can be calculated in $\mathcal{O}\left(|V|^2\cdot A^{-1}(|V|^2)\right)$ using the algorithm proposed in~\cite{Edelsbrunner2002} where $A^{-1}$ is the notoriously slowly growing inverse of the Ackermann function~\cite[Chapter~21]{introduction_to_algorithms}, persistent homology in higher dimensions is harder to compute.
The complexity of algorithms for computing persistent homology for dimension greater than or equal to one is $O(n^3)$ if $n$ is the number of simplices of the Vietoris--Rips complex and Gaussian elimination is used to find ranks of matrices of boundary operators, or $O(n^\omega)$ where $\omega$ is the exponent of matrix multiplication (currently $2.3729$) if sparsity of boundary matrices is taken into account, as in \cite{NOmega}. In its turn, the number of simplices $n$ depends cubically on the number $|V|$ of vertices of the functional graph
if persistence diagrams are drawn only in homological dimension one, which requires determination of simplices up to dimension two.

In practice, this limits persistence diagram computations to a few thousand vertices. In~order to alleviate these problems in neural networks with a large set of neurons, we introduce sampling strategies for both the input dataset and the functional graphs.

\medskip

\noindent
\textbf{Sampling the input space.} We compute activation vectors $A_v$ for a fixed subsample $\mathcal{D}'\subseteq\mathcal{D}$. 
In order to justify that this subsampling does not affect the results of the analysis, it is enough to verify that $\text{corr}(A_{v_i}(\mathcal{D}'),A_{v_j}(\mathcal{D}'))$ is sufficiently close to $\text{corr}(A_{v_i}(\mathcal{D}),A_{v_j}(\mathcal{D}))$, and that small variations in the correlation coefficients produce small changes in the persistence diagrams.
This claim is justified by the fact that, if $X$ and $Y$ are random variables with non-null variance and $X^n$ and $Y^n$ denote sequences of $n$ samples from $X$ and $Y$ respectively, then the sample correlation of $X^n$ and $Y^n$ converges in probability to the correlation between $X$ and $Y$ by the law of large numbers and the continuous mapping theorem~\cite{Statistics}.

In practice, $\mathcal{D}'$ is fixed to a uniform sample of $2{,}000$ elements from the original training dataset, an experimentally selected size that is large enough to obtain sufficient precision.

\medskip

\noindent
\textbf{Sampling the functional graph.}
Because of computational limitations, in the case of modern DNNs less than $1\%$ of the nodes ---a priori, a statistically insignificant sample size--- can be included in the persistent homology calculation. To alleviate this, we sample nodes according to a notion of importance,
following ideas introduced in~\cite{adaptiveHierarchicalDownSampling} adapted to neurons on a neural network instead of inputs of the dataset.
Thus, let $\mathcal{D}'$ be some selected subsample of the training dataset. The \emph{importance score} of a node $v\in V$ is defined as
\begin{equation}
\label{importance}
    I_v(\mathcal{D}') =
    \left|\left\{x\in\mathcal{D}' : v = \argmax_{w\in V}\left|N_w(x)\right|\right\}\right|,
\end{equation}
where $\argmax$ returns only one vertex in case of tie between multiple vertices
---in our case, we use the tie breaking strategy implemented by the NumPy library~\cite{harris2020array}.
Hence $I_v(\mathcal{D}')$ indicates the amount of inputs from $\mathcal{D}'$ for which the activation of $v$ is the largest (or tied-to-largest) among all nodes. Note that a majority of nodes $v$ will have $I_v(\mathcal{D}') = 0$. This is equivalent to excluding these nodes from analysis, which is undesirable ---not only because it is unclear how this will affect the application of TDA, but also because the amount of nodes with $I_v(\mathcal{D}') \neq 0$ might be low enough to severely constrain the size of a subsample. Thus, from $I$ we construct a probability distribution $P$ on~$V$, artificially inflated to make sure that every element of $V$ appears with nonzero probability. This probability $P(v)$ is defined as 
\begin{equation}
\label{probability}
\frac{I_v(\mathcal{D}')}{|\mathcal{D}'|+1} 
\;\text{
if $I_v(\mathcal{D}')>0$, and}
\;\;
\frac{1}{(\left|\mathcal{D}'\right| + 1)\cdot{\left|\{u\in V: I_u(\mathcal{D}') = 0\} \right|}} \;\;
\text{otherwise.}
\end{equation}
Specifically, we sample $3{,}000$ nodes (without repetition) according to this probability distribution, and restrict our analysis to these nodes. This sampling is non-deterministic, and thus can be repeated a number of times to obtain $n$ different subsamples $V_1,\dots,V_n$. Applying the same transformations on the $n$ resulting functional graphs we obtain $n$ different persistence diagrams  per network. Then, we use bootstrapping over the $n$ summaries (see \ref{methodTopoSum}) combining them into a single one. This last representation aims to approximate the persistence summary that would be obtained without sampling.

\begin{table}[ht]
\centering
\caption{Top three combinations of persistence summaries per task according to their respective mean of $R^2$ test values in the $10$ experiments of the $5\times 2$-fold cross-validation statistical test.
\textbf{ASD:} Average and standard deviation of births and deaths.
\textbf{ASDSQ:} Average and standard deviation of births and deaths, concatenated with the corresponding squared values; see Section~\ref{sec:experimentsMain}.}
\label{tab:bestTDASummaries}
\begin{tabular}{@{}ccc@{}} \toprule
\multicolumn{3}{c}{Task 1} \\ \midrule
Top TDA summaries & Best dim & $R^2$ score
\\ \midrule
ASDSQ
& 0 and 1 & $0.5601 \pm 0.13$ \\
ASDSQ& 1 & $0.4321 \pm 0.12$ \\
ASD
& 1 & $0.3720 \pm 0.14$ \\ \midrule

\multicolumn{3}{c}{Task 2} \\ \midrule
Top TDA summaries & Best dim & $R^2$ score \\ \midrule
ASD & 1 & $0.9337 \pm  0.01$\\
ASD & 0 and 1 & $0.9198 \pm 0.02$\\
ASDSQ & 1 & $0.9166 \pm  0.03$\\ \bottomrule
\end{tabular}
\end{table}

\subsection{Discussion}
\label{sec:disc}

\begin{figure*}[htb]
\centering
\begin{subfigure}{\textwidth}
  \centering
\includegraphics[width=\textwidth]{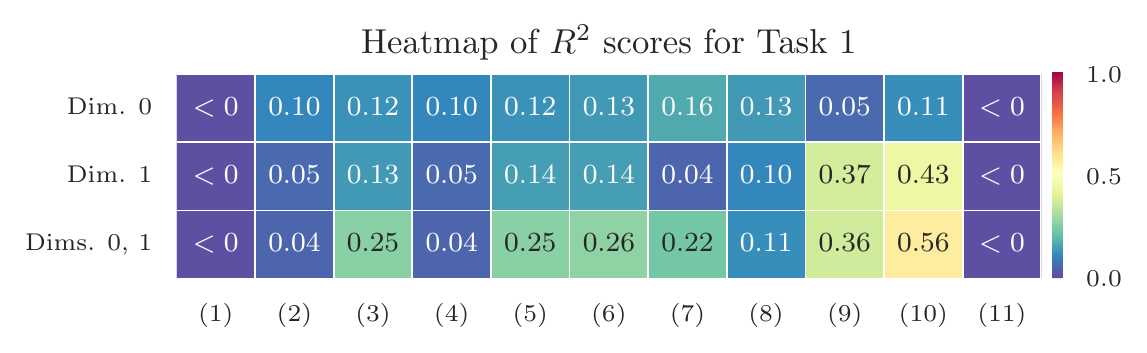}
\label{fig:task1heatmap}
\vspace{-1.3\baselineskip}
\end{subfigure}
\begin{subfigure}{\textwidth}
\centering
\includegraphics[width=\textwidth]{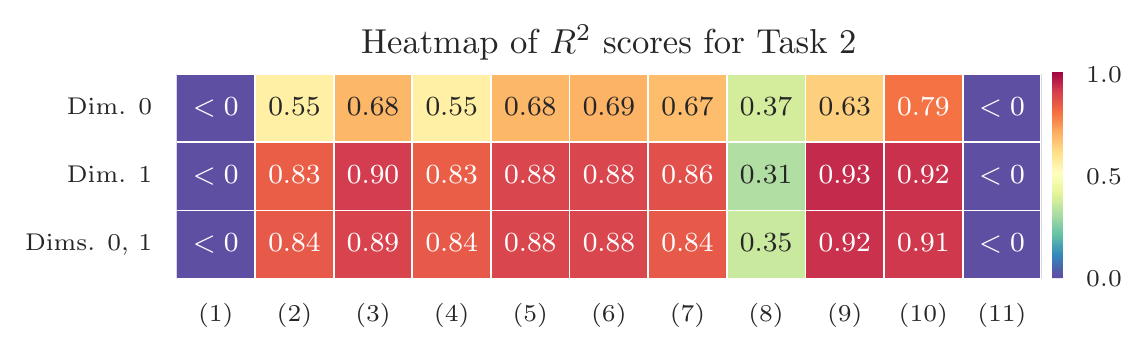}
\label{fig:task2heatmap}
\end{subfigure}
\caption{Mean $R^2$ test values after the 10 experiments of the $5\times 2$-fold cross-validation statistical test for tasks $1$ and $2$ for the combinations of persistence summaries described in the experimental procedure of Section~\ref{sec:experimentalSetup}.  
Rows correspond to homological dimensions $H_0$, $H_1$, and a concatenation of both.
Column numbers represent the following combinations of persistence summaries:
(1)~persistence pooling of $10$ elements; 
(2)~average lives and midlives; (3)~average lives and midlives, original and squared;
(4)~average births and deaths; (5)~average births and deaths, original and squared; (6)~average births and deaths with a logarithmic model; (7)~concatenation of combinations 3 and 5; (8)~persistence entropy; (9)~average and standard deviation of births and deaths; (10)~average and standard deviation of births and deaths, original and squared; ($11$)~complex polynomials with $10$ coefficients.}
\label{fig:ourHeatmaps}
\end{figure*}

\noindent
The combinations of persistence summaries that yielded the top three mean $R^2$ scores for the generalization gap prediction experiments are shown in Table~\ref{tab:bestTDASummaries}. Basic statistical descriptors related to births and deaths of homology generators obtained highest scores overall, validating the results obtained in~\cite{ali2022survey}, in which simple vectorizations consisting of elementary statistical descriptors of persistence diagrams were the persistence summaries that obtained the best performances as input in a variety of image classification tasks. In particular, the vectors composed of averages and standard deviations of births and deaths (and their squares) were those that obtained the best $R^2$ scores in both tasks.
Figure\;\ref{fig:ourHeatmaps} 
shows the average performance of the entire list of summaries.
These results suggest that the generalization gap is mostly linked with the average position and dispersion of points in persistence diagrams. Summaries based on alleged predominance of larger lifetime values, such as persistence entropy or persistence pooling vectors, showed a lower predictive value.

Overall, results are more conclusive for Task~2 than for Task~1, and more significant in homological dimension~$1$, although some of the best $R^2$ scores are achieved using a combination of dimensions $0$ and $1$ for both tasks.
It should also be noticed that $R^2$ scores grow when squares of summaries are added to the model, suggesting departure from linearity.

\medskip

\noindent
\textbf{Explainability.}
\label{sec:explainability}
The distribution of points in persistence diagrams is determined by correlations between neuron activation vectors.
Generators of the zero-homology group $H_0$ of a Vietoris--Rips simplicial complex at filtration level $t$ correspond to connected components of a functional graph in which every edge has a weight smaller than or equal to~$t$, hence a correlation coefficient of $1-t$ in absolute value among the neurons in the group. Hence, for $t=0$ there is one generator for each group of neurons that share correlation coefficients equal to~$\pm 1$.
Points $(0,d)$ in zero-dimensional persistence diagrams arise whenever two (or more) connected components merge in the filtration at time~$d$, and therefore they correspond to non-zero edge weights of a minimum spanning tree of the network's functional graph. 
High weights in a minimum spanning tree imply that the overall correlations between neurons are low.
The lower the correlation between neurons, the higher the number of nonlinearly related features learned by the neural network, and hence the stronger the real expressive power of the network. 
In conclusion, a combination of a high average of death values with a low standard deviation in a zero-dimensional persistence diagram is a plausible indication of an increased expressive power of the neural network, that should lead to better generalization capabilities and thus a smaller generalization gap.

Points in one-dimensional persistence diagrams correspond to cycles of the network's functional graph that are not filled by regions in the Vietoris--Rips complex. Thus a one-dimensional generator appears in the filtration at time $t$ whenever there is a cyclically ordered group of neurons sharing correlations greater than or equal to $1-t$ with their neighbours, which can be interpreted as a group of neurons that have learned similar features.
The earlier a cycle is born, the higher the correlations among the neurons in the cycle, and the higher the death value of a cycle, the higher the differences between the features learned by non-neighbouring neurons in the cycle. Therefore, higher lifetime values
may be associated with an increased number of different features learned by groups of jointly operating neurons.
Thus, the higher the deaths in the one-dimensional persistence diagram of the functional graph of a neural network, the more expressive power the neural network may have, and thus the better it may generalize. 

\medskip

\begin{figure*}[htb]
\centering
\includegraphics[width=\textwidth]{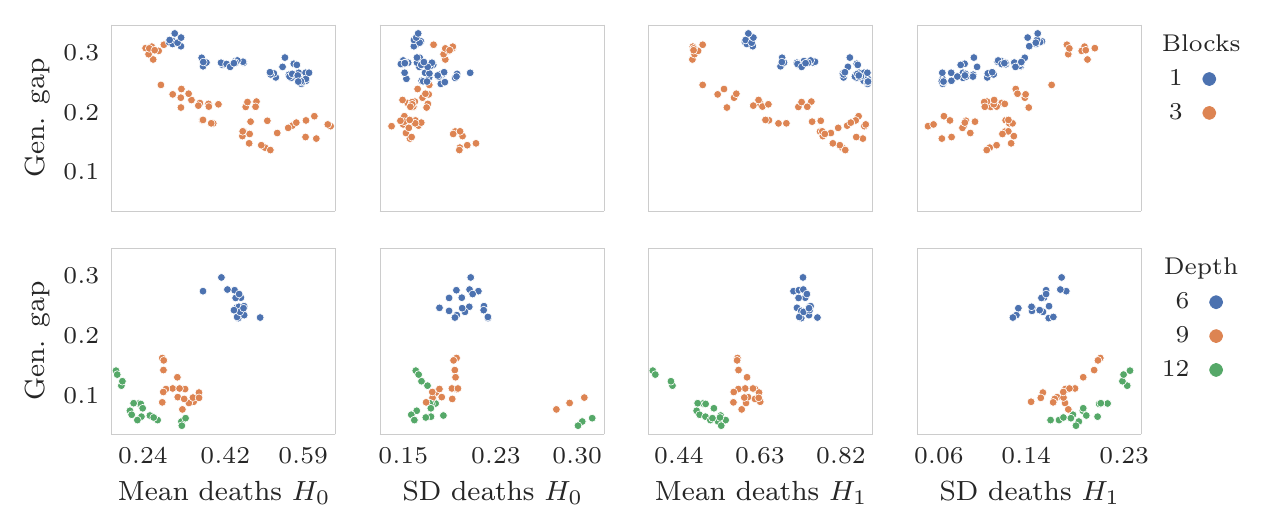}
\caption{Averages and standard deviations of deaths for persistence diagrams in dimension $0$ (first two columns) and dimension $1$ (last two columns) for Task~1 (first row) and Task~2 (second row). For Task~1, points represent 96 VGG-like neural networks trained on the CIFAR10 dataset;
blue and orange points represent neural networks with one and three convolutional blocks, respectively.
For Task~2, points correspond to 54  \emph{network in network} architectures trained on the SVHN dataset; 
blue, orange, and green points represent neural networks with six, nine, and twelve convolutional layers, respectively.}
\label{fig:avgstdFeature}
\end{figure*}

\noindent
\textbf{Clustering.}
\label{sec:clustering}
The interpretations described in the previous subsection are consistent with what is shown in Figure~\ref{fig:avgstdFeature}. In this figure, each row represents a different task, each column represents a different persistent summary, and each point in a cell corresponds to a neural network for the given task. 
The two rows, upper and lower, represent Task~1 and Task~2, respectively. The first and third columns represent the average of deaths of zero- and one-dimensional persistence diagrams, whereas the second and fourth columns represent the standard deviation of deaths of zero- and one-dimensional persistence diagrams, respectively. In the first and second rows, neural networks are clustered according to the number of convolutional blocks and the number of convolutional layers that each network contains.

Figure~\ref{fig:avgstdFeature} suggests that persistence summaries detect very neatly the clusters of neural networks in each task. Naturally, the generalization gap is strongly influenced by the depth of the networks, which is almost determined by the number of convolution blocks and layers. 
When the number of convolutions is fixed, we see a consistent behavior:
the higher the average deaths and the lower the standard deviations, the better the network's performance.
This discovery has the potential of being used for network regularization.

We further analyzed if persistence diagrams for individual labels in a classification task were different between them, in order to gain insight about what was influencing TDA methods and functional graphs
the most. We computed persistence diagrams in dimensions $0$ and $1$ per different neural network and per label. The datasets used to recreate functional graphs were restrictions of the test set to each label.
Details and figures can be found in the Appendix.
Similar results were seen when comparing these persistence diagrams with the original ones. The majority of class-dependent persistence diagrams whose DNNs obtained extreme accuracies, i.e., highest and lowest, were analogous to the diagrams in the class-independent case. This shows that functional graphs are robust to unbalanced datasets in terms of the number of samples per label. 

\begin{table}[ht]
\centering
\caption{Comparison of our best performing summaries with state of the art:
Average and standard deviation of $R^2$ scores for Task~$1$ and Task~$2$ computed from linear models trained in the ten cases of the $5\times 2$-fold cross-validation.}
\label{fig:winnerandbesttable}
\begin{tabular}{@{}lrr@{}}
\toprule
& Task 1 
\hspace*{1cm}
& Task 2 
\hspace*{1cm}
\\ \midrule
Interpex          & $-0.0518 \pm 0.06$\hspace*{0.5cm}
& $0.9500\pm 0.01$\hspace*{0.5cm} 
\\
Always Generalize & $0.9715 \pm 0.01$\hspace*{0.5cm} 
& $0.8893 \pm 0.02$\hspace*{0.5cm}
\\
BrAIn             & $0.4520 \pm 0.08$\hspace*{0.5cm}
& $0.7180 \pm 0.04$\hspace*{0.5cm}
\\ \midrule
Ours  & $0.5601 \pm 0.13$\hspace*{0.5cm}
& $0.9337\pm 0.01$\hspace*{0.5cm}
\\ 
\end{tabular}
\end{table}

\begin{table}[ht]
\centering
\caption{Statistical $p$-values of the pairwise $5\times 2$-fold cross-validation significance test proposed in~\cite{statistical_tests}, with the coefficient of determination $R^2$ as performance metric. The null hypothesis is that, for a fixed-size random drawn training dataset, the two linear models trained with our combinations of persistence summaries or with the winning generalization measures of the PGDL competition have the same $R^2$ score on a randomly drawn test dataset. 
Boldface $p$-values are lower than $0.05$. 
\textbf{ASD1:} Average and standard deviation of births and deaths of dimension one.
\textbf{ASDSQ01:} Average and standard deviation of births and deaths, concatenated with their squared values, for dimensions zero and one.
Their $R^2$ scores are shown in Table~\ref{tab:bestTDASummaries}.}
\label{fig:significance_test}
\begin{tabular}{@{}ccccc@{}}
\toprule Task~1
& Interpex
& Always Generalize
& BrAIn
& ASDSQ01
\\ \midrule      

ASDSQ01 & 
{\boldmath $0.00$}
& 
{\boldmath $0.01$}
& 
$0.23$
&
\\ 
ASD1 &  
{\boldmath $0.03$}
& 
{\boldmath $0.00$}
& 
$0.55$
& 
$0.13$
\\
\midrule Task~2
&
&
&
&
\\ \midrule  
ASDSQ01 & 
$0.37$
& 
$0.80$
& 
$0.37$
& 
\\ 
ASD1 & 
$0.19$
& 
{\boldmath $0.00$}
& 
{\boldmath $0.01$}
& 
$0.51$
\\ \bottomrule
\end{tabular}
\end{table}

\medskip

\noindent
\textbf{Persistence summaries.} Results show that linear models of persistence summaries can predict the generalization gap, since we obtained competitive results in both tasks, as seen in Table~\ref{tab:bestTDASummaries} and Table~\ref{fig:winnerandbesttable}. However, the fact that a summary based on a combination of non-linear transformations of persistence features yielded the best score for Task~$1$ suggests that more complex models can have better capacity to relate persistence summaries to the generalization gap. 

When it comes to ranking summaries, persistence pooling and complex polynomials produced the lowest $R^2$ scores overall, as shown in Fig.~\ref{fig:ourHeatmaps}.
For persistence pooling, one possible explanation of its low performance is that it relies on lifetimes of points that live the longest, in contrast to the most effective summaries, which are based on average location and dispersion of the whole set of points in a persistence diagram. Similarly, truncated complex polynomials are not sufficiently accurate measures of the location and aggregation of the collection of all points in persistence diagrams. The fact that persistence entropy achieves non-optimal $R^2$ scores for Task~2 in Fig.~\ref{fig:ourHeatmaps}
is consistent with the interpretation that the distribution of points near the diagonal in one-dimensional persistence diagrams is substantial for generalization gap prediction.

\medskip

\noindent
\textbf{State-of-the-art comparison.}
Table~\ref{fig:winnerandbesttable} shows a comparison of the results of our best performing linear models based on persistence summaries with state-of-the-art methods. In this table, the $R^2$ scores describe the ability of each linear model
to predict the generalization gap with respect to the coefficient of determination.
Table~\ref{fig:significance_test} shows the pairwise $p$-values between the linear models induced by our best performing combinations of persistence summaries, shown in Table~\ref{tab:bestTDASummaries}, and the linear models induced by the winning generalization measures of the PGDL dataset.

We obtain the second-best mean $R^2$ scores for both tasks, after Always Generalize and Interpex in the first and second ones, respectively. However, assuming that two methods are significantly different whenever their pairwise $p$-value is lower than $0.05$ in Table~\ref{fig:significance_test}, there is no significant difference between the $R^2$ scores of the linear models of our best combination of persistence summaries in Task~2 and the linear models induced by the generalization measure of the Interpex team. Additionally, our models are significantly better than those for the Interpex generalization measure in Task~1 and than the ones for the generalization measures of Always Generalize and BrAIn in Task~2. These results suggest that persistence summaries are a promising tool to develop robust models to predict the generalization gap.

\subsection{Hardware, software and licenses}
\label{sec:HardwareSetups}

\noindent
Persistence diagrams were computed with Python \texttt{giotto-ph}~\cite{burella2021giottoph} (GNU AGPLv3) using a Quadro P6000 GPU. Persistence summaries were computed with the \texttt{giotto-tda} framework ~\cite{tauzin2020giottotda} 
(AGPLv3 License), and density curves were drawn using SciPy 1.8.0~\cite{2020SciPy-NMeth}. Analysis was performed on a personal computer with an Intel Core i7 (4th~generation) processor with an NVIDIA GeForce GTX 960M 2GB GDDR5, using the libraries Jupyter Notebook (New BSD License), NumPy (BSD 3-Clause ``New'' or ``Revised'' License) and TensorFlow with Keras (Apache 2.0 License).
Docker (Apache 2.0 License) was also used to perform the experiments.
The dataset of neural networks from~\cite{generalizationChallenge} is licensed under Apache 2.0.

\section{Conclusions}
\label{sec:conclusions}

\noindent
We have defined a framework that can be used to explore interpretability of DNNs based on topological properties of their functional graphs. This relaxes the problem of understanding the internal representations of a neural network to, in a broad sense, understanding their \textit{shape}. Regarding generalization,
we have shown examples of how one can interpret DNN neuron interactions based on their correlations by means of persistence diagrams. Moreover, we proved that the generalization gap can be consistently predicted using topological persistence summaries extracted from 
functional graphs,
with a competitive prediction accuracy on two different computer vision problems. The most successful summaries were those related with the average location and dispersion of points in persistence diagrams. Hence, it is not true in our case that points near the diagonal in persistence diagrams are irrelevant, as often claimed in TDA studies.

\medskip

\noindent
\textbf{Limitations.}
\label{sec:limitations}
A practical limitation of persistent homology comes from its computational complexity ---sampling methods are not necessarily optimal and information might be lost in sampling processes for datasets and for neurons. Transformations of persistence diagrams into summaries may also cause a loss of information; however, this seems unavoidable if one wants to obtain easy-to-compute generalization measures.

\medskip

\noindent
\textbf{Future work.}
Although we found strong patterns relating persistence summaries with generalization gaps (Figure~\ref{fig:avgstdFeature}), broader experimentation is required to see if these patterns are consistent among other kinds of networks and machine learning tasks, and also to make more explicit which features of the networks are involved in the TDA-driven clustering effect that we have observed.

The mere definition of functional graphs raises a question: which is the optimal metric to compare neurons given an architecture? There might be better alternatives to linear correlation between activation vectors; for instance, Spearman correlation was used in co-activation graphs for a similar purpose in~\cite{Horta}. 

Another problem is to find an optimal neuron sampling strategy. This is related to the problem of finding the most relevant neurons in a DNN graph. Persistence summaries suggest that grouping neurons in terms of their activation structure is feasible for DNNs. However, understanding which functional phenomena are being captured into such communities of nodes needs further study. This could lead to the discovery of new architectural properties useful to develop better networks. 

Figure~\ref{fig:avgstdFeature} shows that, fixing the depth of a neural network, there is a consistent association between a lower generalization gap and a higher average of death values together with a small dispersion in the persistence diagrams of the network's functional graph in dimensions zero and one.
This finding has the potential to improve the performance of a given architecture during training by means of a regularization term that maximizes averages of deaths while minimizing standard deviation, using the framework for differential calculus  peronsistence diagrams discussed in~\cite{Leygonie2022, pmlr-v139-carriere21a}.

\bibliographystyle{elsarticle-num.bst}
\bibliography{refs}

\section*{Appendix}

\begin{figure*}[t]
  \centering
\includegraphics[width=\textwidth]{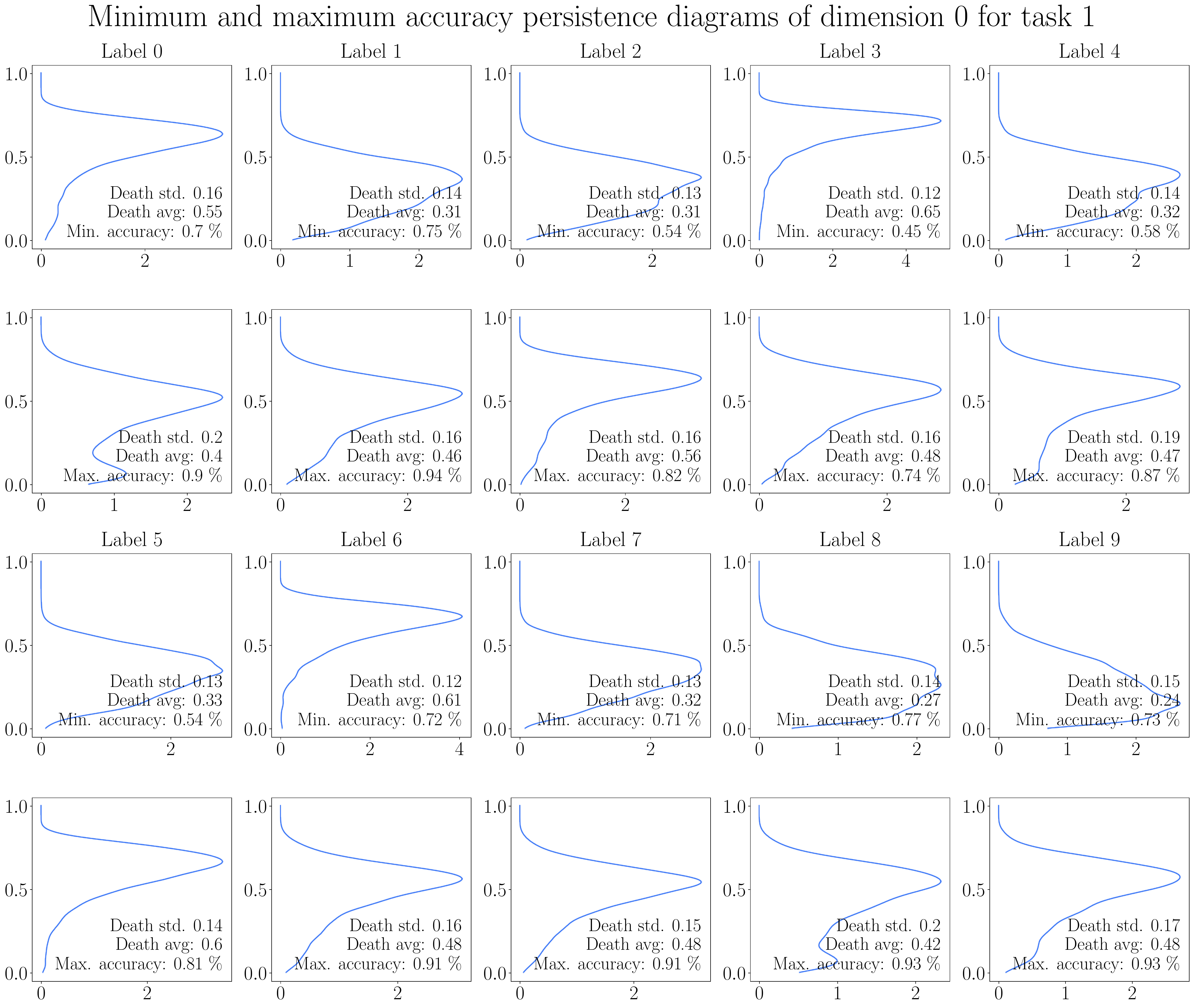}
\caption{Lifetime densities in
persistence diagrams 
in homological dimension zero of $96$ VGG-like neural networks with minimum and maximum accuracies on the test set per label for Task~$1$.}
\label{fig:individualLabelsTask1Dim0}
\end{figure*}

\begin{figure*}[t]
\centering
\includegraphics[width=\textwidth]{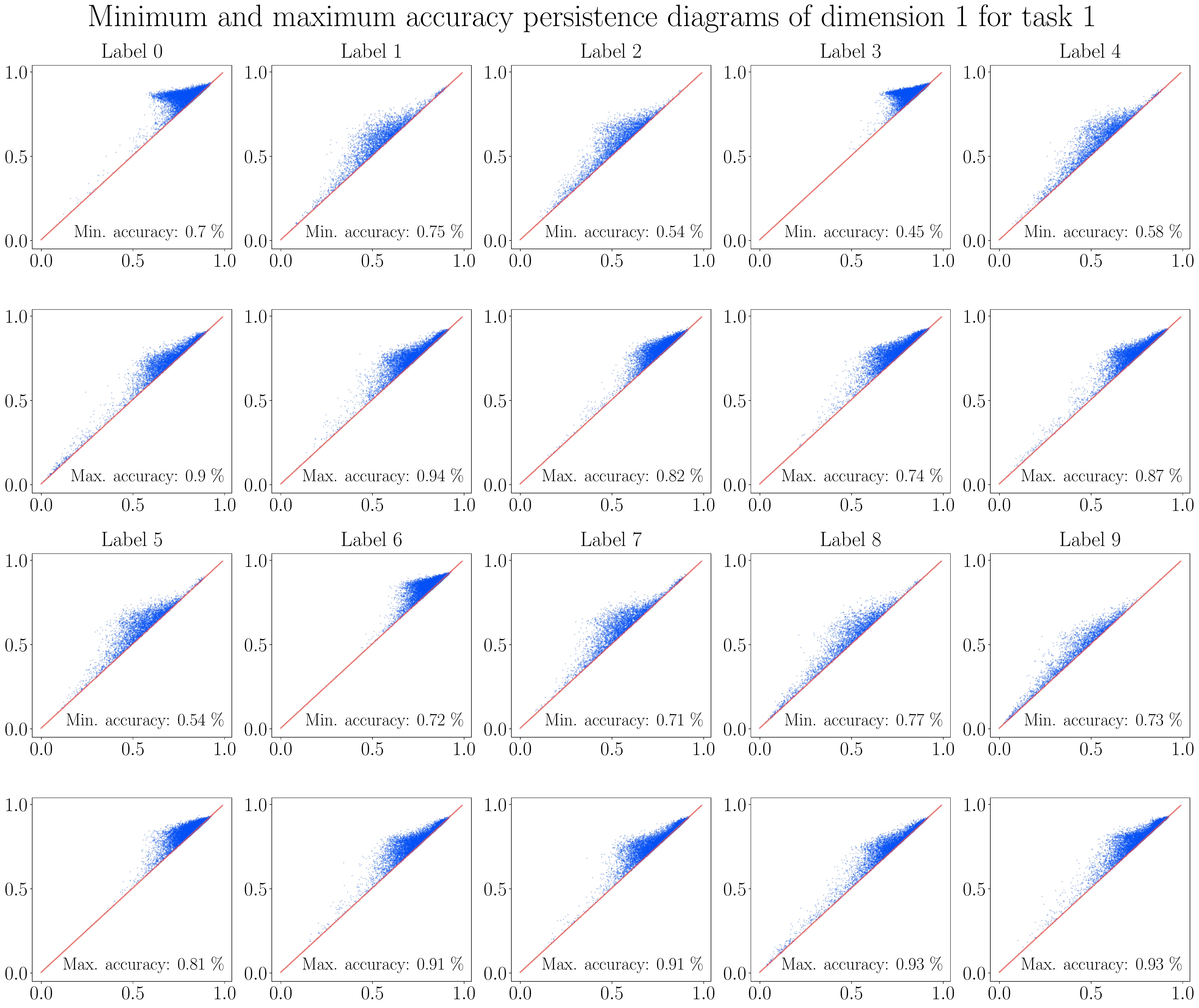}
\caption{Persistence diagrams in homological dimension one of 96 VGG-like neural networks with minimum and maximum accuracies on the test set per label for Task~$1$.}
\label{fig:individualLabelsTask1Dim1}
\end{figure*}

\begin{figure*}[t]
  \centering
\includegraphics[width=\textwidth]{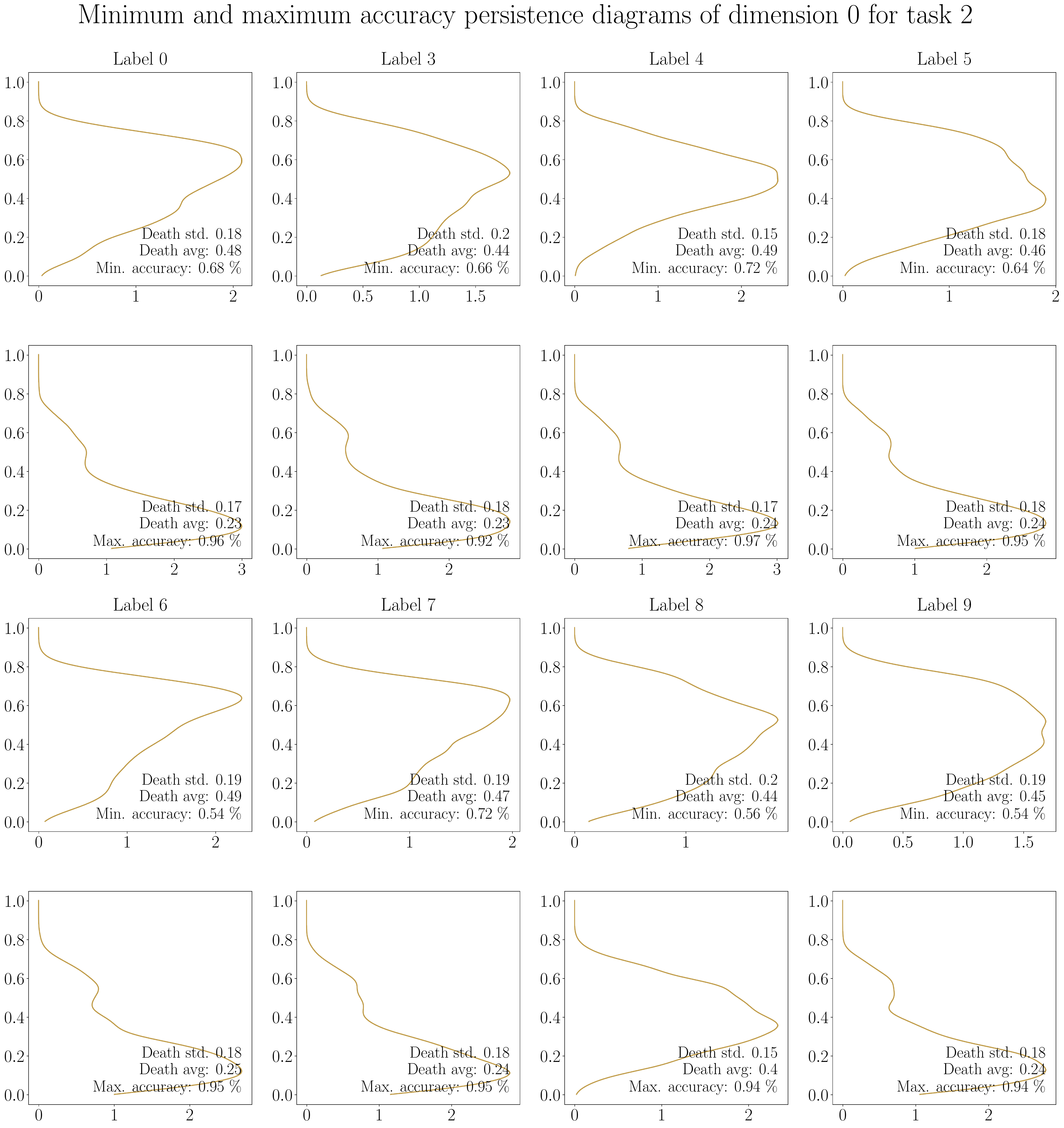}
\caption{Lifetime densities in
persistence diagrams 
in homological dimension zero of $54$ \emph{network in network} architectures with minimum and maximum accuracies on the test set per label for Task~$2$.}
\label{fig:individualLabelsTask2Dim0}
\end{figure*}

\begin{figure*}[t]
\centering
\includegraphics[width=\textwidth]{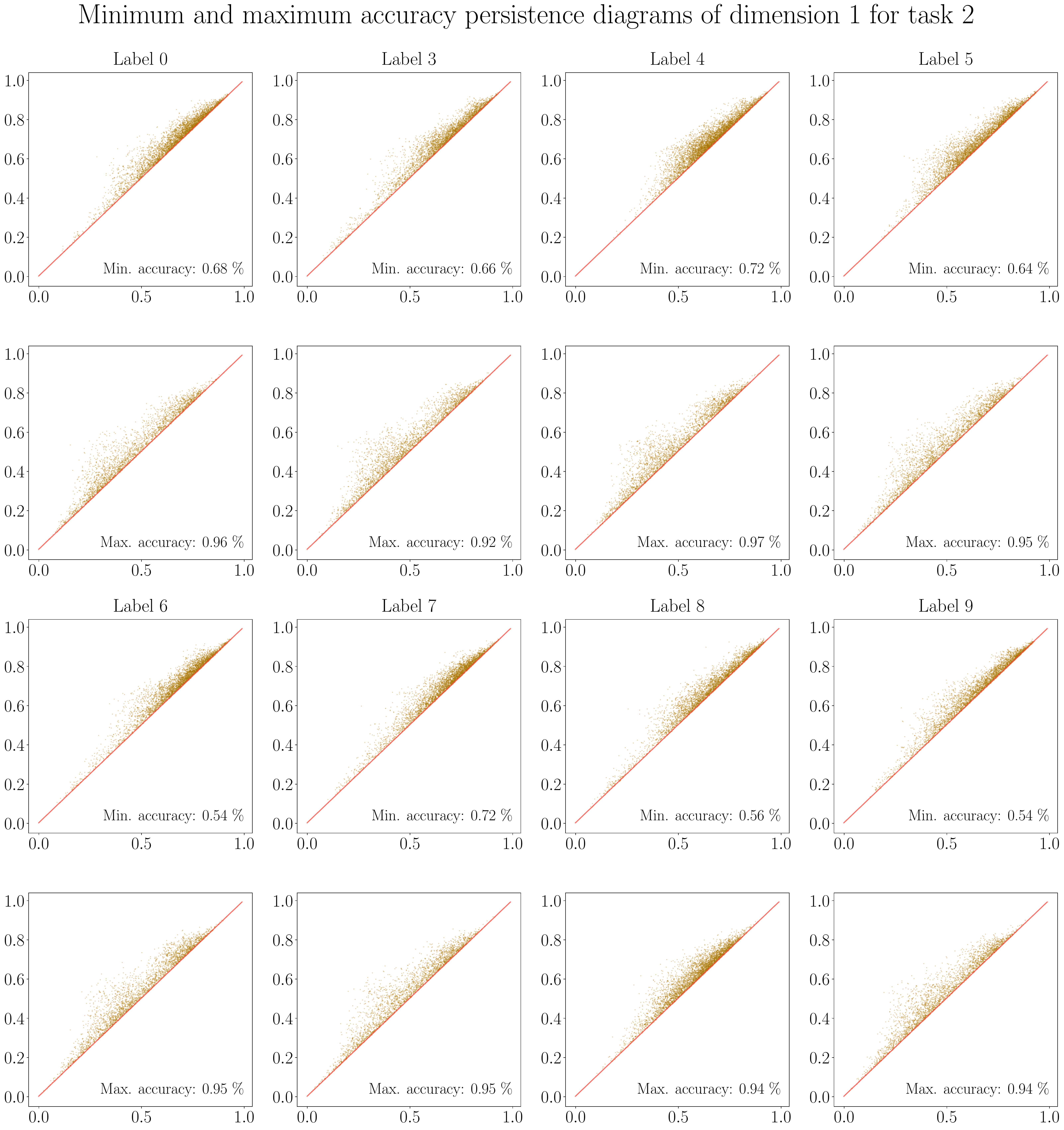}
\caption{Persistence diagrams in homological dimension one of 54 \emph{network in network} architectures with minimum and maximum accuracies on the test set per label for Task~$2$.}
\label{fig:individualLabelsTask2Dim1}
\end{figure*}

\noindent
This appendix contains an analysis per label of persistence diagrams in dimensions $0$ and~$1$. 
The datasets that we used to recreate 
functional graphs were the restriction of the test sets to each label. We computed accuracy for each of these test subsets, and plotted persistence diagrams 
corresponding to those neural networks that achieved the maximum and minimum accuracies on 
test subsets per label for dimensions $0$ and~$1$.
The results can be seen in Figures\;\ref{fig:individualLabelsTask1Dim0}, \ref{fig:individualLabelsTask1Dim1}, \ref{fig:individualLabelsTask2Dim0} and \ref{fig:individualLabelsTask2Dim1}.
These results are consistent with what we found in persistence diagrams computed with the whole training dataset. 
Thus we see that distinction between inputs of different labels does not have a substantial influence on the distribution of points in persistence diagrams.

For a more convenient visualization, persistence diagrams in dimension~$0$
have been replaced with lifetime density curves, calculated by means of Gaussian kernels.
Lifetime values are equal to death values for zero-homology generators.

It can be seen in
Fig.~\ref{fig:individualLabelsTask2Dim0}
and
Fig.~\ref{fig:individualLabelsTask2Dim1} that increased accuracy values for Task~2 match with scattering of points downwards the diagonal of the persistence diagram in dimension~$1$ and with a lower average life in dimension~$0$. 
This pattern is apparently not consistent with other architectures, such as those used in Task~1. This is explained by the splitting of network types into clusters as observed in Fig.~\ref{fig:avgstdFeature}, since for Task~2 the regression line for average deaths has negative slope in each cluster, while it has positive slope if clustering is not taken into account.

\end{document}